\documentclass[conference]{IEEEtran}
\IEEEoverridecommandlockouts
\IEEEaftertitletext{\vspace{-9pt}}
\usepackage[T1]{fontenc}
\usepackage{cite}
\usepackage{amsmath,amssymb,amsfonts}
\usepackage{graphicx}
\usepackage{textcomp}
\usepackage{booktabs,array,tabularx}
\usepackage{balance}
\usepackage[hidelinks]{hyperref}
\graphicspath{{figures/}}
\newcolumntype{Y}{>{\raggedright\arraybackslash}X}

\newcommand{\FQ}{\mathrm{FQ}}
\makeatletter
\newcommand{\linebreakand}{%
  \end{@IEEEauthorhalign}\hfill\mbox{}\par
  \mbox{}\hfill\begin{@IEEEauthorhalign}}
\makeatother

\begin{document}

\title{SearchInject: Retrieval Exposure Control\\
for Indirect Prompt Injection in\\Language-Model Search}

\author{
\IEEEauthorblockN{Kainan Zhou}
\IEEEauthorblockA{\textit{Google LLC}\\Mountain View, USA\\zhoumark@google.com}
\and
\IEEEauthorblockN{Gangzhen Qian}
\IEEEauthorblockA{\textit{Google LLC}\\Mountain View, USA\\irisqian@google.com}
\and
\IEEEauthorblockN{Janet Sung}
\IEEEauthorblockA{\textit{Google LLC}\\Mountain View, USA\\janetsung@google.com}
\linebreakand
\IEEEauthorblockN{Zhi Jing}
\IEEEauthorblockA{\textit{Google LLC}\\Mountain View, USA\\varusjing@google.com}
\and
\IEEEauthorblockN{Rong Bao}
\IEEEauthorblockA{\textit{Google LLC}\\Mountain View, USA\\arielbao@google.com}
\and
\IEEEauthorblockN{Qianru Lao}
\IEEEauthorblockA{\textit{Harvard University}\\Cambridge, USA\\estherbear17@gmail.com}
}
\maketitle

\begin{abstract}
An indirect prompt injection can affect a search-augmented large language model only after its carrier page reaches model context. SearchInject controls this admission decision with a joint score combining retrieval relevance, search-abuse cues, instruction likelihood, and two relevance interactions. A quarantine threshold is selected under an empirical 0.5\% clean-candidate false-quarantine constraint on a calibration split. Evaluation uses 300 held-out SciFact queries and held-out payload categories within four construction families already represented during development. Across 1,200 query--family outcomes per retriever, top-10 target exposure falls from 0.9408 to 0.1583 for BM25, from 0.9158 to 0.1508 for a normalized hybrid, and from 0.8450 to 0.1375 for latent semantic analysis. These are relative reductions of 83.2--83.7\%; the largest absolute change in clean normalized discounted cumulative gain at rank 10 is 0.0018. The frozen threshold yields clean test false-quarantine rates of 0.59--0.65\%, above the empirical calibration constraint. A query-level analytical reanalysis accounts for the repeated construction outcomes under an independent-query assumption. A separate 104-case generation follow-up records context removal but no confirmed undefended hijacks under a post hoc endpoint, so it cannot estimate a reduction in attack success. The evidence supports retrieval exposure control within the evaluated construction families; the benefit of the interaction terms at matched clean cost remains unresolved.
\end{abstract}

\begin{IEEEkeywords}
indirect prompt injection, information retrieval, retrieval-augmented generation, language models
\end{IEEEkeywords}

\section{Introduction}
Search-augmented language models answer queries using retrieved third-party text. Publishers can embed instructions without authority over the application; early demonstrations showed that such content could redirect language-model applications~\cite{greshake}. In search, the carrier page must first be retrieved, ranked, and admitted to context.

Query-term repetition, title manipulation, doorway text, and copied passages can promote a page~\cite{gyongyi}. When results become model inputs, promotion determines which untrusted instructions the generator encounters, creating an admission problem before any instruction is followed.

InjecAgent and AgentDojo evaluate behavior after hostile content is delivered through tool observations or task environments~\cite{injecagent,agentdojo}. Their endpoints do not isolate carrier selection.

Retrievable trigger fragments can cross this selection boundary under ordinary-looking queries~\cite{retrievalbarrier}. SearchInject addresses the admission decision and its clean retrieval cost.

The main technical contribution is a relevance-conditioned risk score coupled with a clean-cost-constrained quarantine threshold. Logistic scoring and refill implement this admission policy. The evaluation separates target exposure from task hijacking, so a change in one quantity is not credited as a change in the other.

Exposure decreases by 83.2--83.7\% across three lightweight retrievers, with small clean ranking changes. We report calibration transfer and query-level uncertainty; interaction effects and matched-cost comparisons remain unresolved.

\section{Background and Related Work}
\subsection{Injection and retrieval}
BIPIA embeds instructions in external content paired with benign tasks~\cite{bipia}. The formulation of Liu et al. treats prompt injection as conflict between the application's objective and an injected objective~\cite{formalizing}. Our exposure endpoint records whether the crafted target page reaches context; a confirmed hijack additionally requires observable execution of the injected task.

Retrieval-augmented generation supplies selected external evidence to a generator~\cite{rag}. PoisonedRAG optimizes inserted content for both retrieval and a chosen answer~\cite{poisonedrag}. SearchInject examines the admission decision and its clean retrieval cost using fixed, attributable page constructions. This setting permits controlled comparisons but does not represent adaptive optimization against the defense.

The constructions modify title coverage, term placement, borrowed evidence, or indexed metadata. These cues connect to content-based spam detection~\cite{ntoulas}. The source-overlap signal is a controlled proxy; reliable ownership and provenance are not available to the defender in this experiment.

\subsection{Complementary defense boundaries}
DataSentinel evaluates detection against a strategically adapting injector, whereas PromptLocate identifies contaminated spans within a document~\cite{datasentinel,promptlocate}. SearchInject scores complete retrieved candidates before they enter the generator context. It does not localize an instruction or certify that an admitted page is safe.

StruQ combines structured queries with model training; SecAlign uses preference optimization~\cite{struq,secalign}. Both strengthen the boundary between trusted instructions and external data after retrieval.

Third-party chatbot plugins introduce additional trust-boundary failures through scraping and application integration~\cite{chatbotweb}. Our fixed corpus does not reproduce those deployment paths.

\section{SearchInject Design}
\subsection{Threat model and admission policy}
The attacker controls a published page's title, body, and markup-like metadata for a known query family. The service controls the retrieval and generation pipeline. The defender observes queries, candidate text, and ranker scores. Pages are fixed before final evaluation; attackers do not adapt to defense outputs.

\begin{figure}[t]
\centering
\includegraphics[width=\columnwidth]{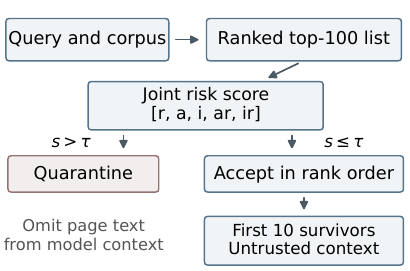}
\caption{SearchInject scores the depth-100 candidate list before context assembly. Rejected pages are omitted, and accepted candidates retain their original relative order. The main evaluation measures target admission; generation behavior is a separate exploratory endpoint.}
\label{fig:pipeline}
\end{figure}

Figure~\ref{fig:pipeline} locates the intervention. A \emph{crafted target page} carries an injected instruction. \emph{Quarantine} withholds a candidate from generation context without assigning a publisher-level judgment. Exposure and confirmed hijacking are recorded separately.

For each query, the policy scans the original depth-100 ranking. Candidates whose risk score exceeds the threshold are skipped; all others are appended in rank order. The scan stops at ten accepted pages or exhaustion of the candidate list. Thus, removed pages are replaced from the same ranking rather than leaving a shortened top ten. There is no reranking of the survivors and no retrieval beyond the original candidate list.

\subsection{Controlled construction families}
Each target pairs a SciFact query with a deterministically selected source document. \emph{Title stuffing} repeats salient terms in the title and lead sentence. A \emph{doorway} page starts with a query-specific summary and then introduces the payload. A \emph{near duplicate} preserves most of a relevant abstract and appends the instruction. \emph{Hidden metadata} places query terms in indexed fields while leaving the visible body comparatively natural.

Construction records retain family, source identifier, payload category, and content hash. Generation truncation preserves the retrieval-oriented lead and instruction-bearing tail; near duplicates remain distinct from clean sources. All families occur during development and testing.

\subsection{Features and risk score}
Let $r(q,d)$ be normalized ranker relevance. An abuse model produces $a(q,d)$ from title coverage, body term density, repetition, source overlap, and markup indicators. Instruction likelihood $i(d)$ uses inverse-document-frequency-weighted word one- and two-grams and character three- to five-grams. The word features capture short instruction phrases; the character features tolerate small spacing and punctuation changes.

The joint model is class-balanced logistic regression with the fixed feature order
\begin{equation}
 z(q,d)=[r,a,i,ar,ir].
\label{eq:features}
\end{equation}
Its output $s(q,d)=\sigma(b+\boldsymbol{\beta}^{\mathsf T}z(q,d))$ is used as a risk score. The two interactions permit the contributions of abuse and instruction signals to vary with relevance. Construction labels and simulated source markers are excluded. No probability-calibration guarantee is attached to the logistic output.

\subsection{Calibration and implementation scope}
The threshold is selected from calibration scores to maximize crafted-page recall subject to an empirical clean-candidate false-quarantine rate at most 0.005. With $\mathcal{C}_0$ denoting scored clean query--document placements,
\begin{equation}
 \widehat{\FQ}_{\rm cal}(t)=\frac{1}{|\mathcal{C}_0|}
 \sum_{(q,d)\in\mathcal{C}_0}\mathbf{1}\{s(q,d)>t\}.
\label{eq:fq}
\end{equation}
The strict comparison $s>t$ is used for quarantine. A document returned for several queries contributes several placements. The selected threshold is $\tau=0.973853$, with calibration recall 0.804 and false quarantine 0.00498. The feature definitions, fitted model, and threshold are then frozen before test evaluation.

This empirical selection rule does not guarantee a 0.5\% bound on new samples. The feature order and threshold alone cannot reconstruct the trained system; Section~\ref{sec:repro} identifies the parameters and records required for replay.

\section{Experimental Method}
\subsection{Data partitions and retrievers}
The study uses SciFact scientific abstracts and relevance judgments~\cite{scifact} in the BEIR retrieval format~\cite{beir}. This format contains 5,183 documents, 809 training queries, and 300 test queries; its test split corresponds to the original SciFact development split. A deterministic hash assigns 655 training queries to fitting and 154 to calibration. The designated 300-query retrieval test split is held out from model and threshold choices.

Payloads are a filtered subset of BIPIA's text attacks: nine fitting categories, three calibration categories, and 13 retained categories from its test split. These are study-specific subset counts, not BIPIA's complete official splits. Malware, fraud, propaganda, and political-persuasion payloads are excluded. Clean SciFact claims and retrieval-evidence sentences provide negatives.

Queries and payload categories are disjoint across partitions; source-document disjointness is not established. The test measures transfer within known construction families, as summarized in Table~\ref{tab:protocol}.

\begin{table}[t]
\caption{Evaluation protocol and known implementation settings}
\label{tab:protocol}
\centering\small
\begin{tabularx}{\columnwidth}{@{}lY@{}}
\toprule
Item & Setting\\
\midrule
Corpus & 5,183 SciFact abstracts in BEIR format\\
Queries & 655 fitting; 154 calibration; 300 test\\
Payload categories & 9 fitting; 3 calibration; 13 test\\
Construction families & Title stuffing, doorway, near duplicate, hidden metadata; all seen during development\\
Retrieval & BM25, LSA, normalized hybrid; depth 100\\
Joint model & Class-balanced logistic regression; $[r,a,i,ar,ir]$\\
Admission & Quarantine if $s>0.973853$; refill to 10\\
Statistical units & 300 queries; four outcomes per query\\
\bottomrule
\end{tabularx}
\end{table}

BM25 supplies lexical ranking with document-length normalization~\cite{bm25}. Latent semantic analysis (LSA) applies truncated singular-value decomposition to the term--document matrix; the hybrid averages normalized BM25 and LSA scores. Retrieval depth is 100. LSA does not represent Dense Passage Retrieval or current learned encoders~\cite{dpr}.

\subsection{Metrics and query-level uncertainty}
Exposure@$k$ is the fraction of query--family cases in which the crafted target appears among the first $k$ pages admitted to context. Clean utility is evaluated on a separate, unattacked corpus copy. Normalized discounted cumulative gain (NDCG@10) rewards relevant documents near the top; mean reciprocal rank (MRR@10) records the reciprocal rank of the first relevant result; Recall@10 measures recovered judged evidence. False quarantine is measured per clean placement.

The four outcomes for one query are repeated observations. For each retriever, let $U_{qf}$ and $J_{qf}$ indicate exposure without defense and with joint quarantine for query $q$ and family $f$. Define the paired query average
\begin{equation}
 D_q=\frac{1}{4}\sum_{f=1}^{4}(U_{qf}-J_{qf}),\qquad
 \widehat{\Delta}=\frac{1}{300}\sum_{q=1}^{300}D_q.
\label{eq:paired}
\end{equation}
Here $D_q\in[-1,1]$: refill can admit a target that was previously below rank ten. The 1,200 query--family outcomes are therefore not treated as 1,200 independent samples.

Conditional on a fixed corpus, fitted model, and threshold, and treating queries as independent sampling units, Hoeffding's inequality~\cite{hoeffding} gives a conservative marginal 95\% interval for the expected paired reduction:
\begin{equation}
 \widehat{\Delta}\ \pm\ \sqrt{\frac{2\log(40)}{300}}
 =\widehat{\Delta}\ \pm\ 0.15682.
\label{eq:bound}
\end{equation}
This analytical reanalysis uses aggregate counts and permits arbitrary dependence among constructions for one query. It does not cover dependence between queries sharing sources or transfer to unseen families. Intervals are individual per retriever, not simultaneous or obtained by bootstrapping reconstructed records.

Reported clean-metric intervals use 10,000 paired bootstrap samples of whole queries. A sharper exposure analysis requires the original query clusters with all family and policy outcomes. Aggregate counts do not determine their covariance or paired contingency tables; unclustered binomial intervals and aggregate McNemar claims are omitted.

\subsection{Exploratory generation protocol}
The 104-case generation cohort crosses two deterministic selections, four families, and 13 payload categories. Targets must enter the undefended hybrid top five. Clean, attacked, and quarantined retrieval are evaluated under conventional and hardened prompts; the latter labels retrieved material as untrusted.

Qwen3-4B and Phi-4-mini-instruct use four-bit weights and deterministic, temperature-zero decoding. Prompts contain five sources capped at 450 characters each. Answers, limited to 128 tokens, require SUPPORTS, CONTRADICTS, or NOT\_ENOUGH\_INFO plus an explanation. This is a fact-verification follow-up.

Two planned independent human reviewers were unavailable. Automated payload-specific execution checks were fixed after preliminary semantic labels had been inspected, making the endpoint post hoc and non-blinded. Observable execution counts as confirmed hijacking; incomplete or underspecified substitutions remain unresolved. Context removal is evaluated separately.

\section{Results}
\subsection{Retrieval exposure and clean utility}
\begin{figure*}[t]
\centering
\includegraphics[width=0.92\textwidth]{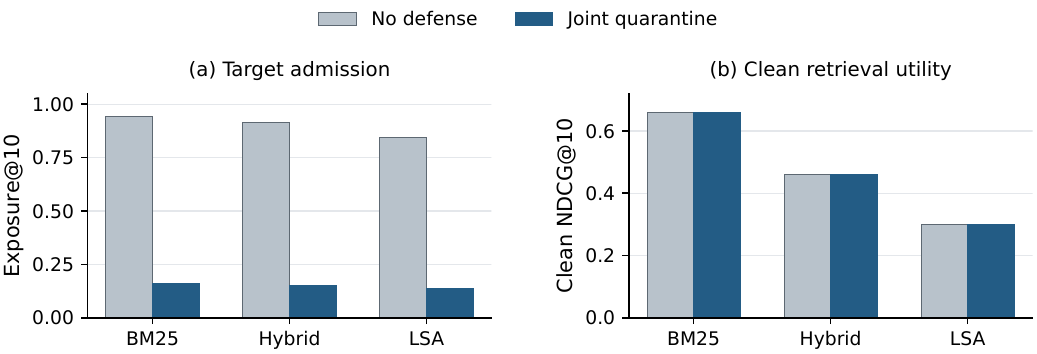}
\caption{Exposure and clean retrieval utility at the recorded operating point. Panel (a) shows the large decrease in target admission; panel (b) shows the small measured clean NDCG change. Bars reproduce Table~\ref{tab:main}; they do not depict an operating-point sweep. The query-level analytical uncertainty for exposure reduction is reported in the text.}
\label{fig:results}
\end{figure*}

Table~\ref{tab:main} and Fig.~\ref{fig:results} summarize the main operating point. Across 1,200 query--family outcomes per retriever, the exact exposed counts change from 1,129 to 190 for BM25, 1,099 to 181 for the hybrid, and 1,014 to 165 for LSA. The absolute reductions are 0.7825, 0.7650, and 0.7075, respectively; relative reductions range from 83.2\% to 83.7\%.

\begin{table}[t]
\caption{Main results over the four known construction families}
\label{tab:main}
\centering\small
\begin{tabular}{@{}lrrrr@{}}
\toprule
 & \multicolumn{2}{c}{Exposure@10} & \multicolumn{2}{c}{Clean NDCG@10}\\
\cmidrule(lr){2-3}\cmidrule(l){4-5}
Retriever & None & Joint & None & Joint\\
\midrule
BM25 & .9408 & .1583 & .6602 & .6584\\
Hybrid & .9158 & .1508 & .4606 & .4593\\
LSA & .8450 & .1375 & .3000 & .2987\\
\bottomrule
\end{tabular}
\end{table}

The conservative 95\% intervals from (\ref{eq:bound}) are [0.626, 0.939] for BM25, [0.608, 0.922] for the hybrid, and [0.551, 0.864] for LSA. These intervals use 300 query units and the independent-query sampling model. The observed benchmark reductions themselves are exact functions of the reported counts.

Clean NDCG@10 changes by $-0.0018$, $-0.0013$, and $-0.0013$ for BM25, hybrid, and LSA. The reported paired bootstrap intervals are $[-0.0053,0]$, $[-0.0039,0.0001]$, and $[-0.0039,0]$. MRR@10 changes by $-0.0017$, 0, and 0; Recall@10 decreases by 0.0017 for each retriever. Small point estimates and intervals near zero do not constitute a formal equivalence test.

\subsection{Calibration transfer and threshold interpretation}
Table~\ref{tab:calibration} separates calibration selection from test observations. The selected rule attains 0.498\% on calibration placements. Frozen-threshold test rates exceed the 0.5\% constraint by 0.09, 0.12, and 0.15 percentage points. Sampling and distribution differences may contribute, but their effects cannot be separated from these rates. Calibration placement counts are unreported, precluding a sampling-error estimate.

\begin{table}[t]
\caption{Empirical calibration constraint and clean test observations}
\label{tab:calibration}
\centering\small
\begin{tabular}{@{}lrr@{}}
\toprule
Split or retriever & False quarantine & Threshold\\
\midrule
Calibration constraint & $\leq 0.500\%$ & Selected on calibration\\
Calibration observed & $0.498\%$ & 0.973853\\
BM25 test & $0.590\%$ & 0.973853\\
Hybrid test & $0.620\%$ & 0.973853\\
LSA test & $0.650\%$ & 0.973853\\
\bottomrule
\end{tabular}
\end{table}

Only one calibrated threshold is reported, so sensitivity cannot be recovered from these aggregates. Lower thresholds increase candidate rejection on a fixed score list, but refill can promote surviving targets. A sensitivity analysis must rerun admission and refill and measure both exposure and clean utility.

\subsection{Construction families and component comparisons}
Table~\ref{tab:families} shows reductions in every construction--retriever pair. Doorway pages have the largest remaining exposure for each retriever, approximately 0.21--0.23. Near duplicates retain exposure under all three retrievers, with lower post-quarantine values for LSA than for BM25 or the hybrid. Copied relevant evidence is a plausible source of difficulty, but these aggregate outcomes do not isolate that mechanism.

\begin{table}[t]
\caption{Exposure@10 by construction family}
\label{tab:families}
\centering\small
\begin{tabular}{@{}lccc@{}}
\toprule
Family & BM25 & Hybrid & LSA\\
\midrule
Title stuffing & 1.000/.083 & 1.000/.083 & .983/.080\\
Doorway & 1.000/.230 & .987/.233 & .847/.207\\
Near duplicate & .763/.197 & .680/.163 & .590/.133\\
Hidden metadata & 1.000/.123 & .997/.123 & .960/.130\\
\bottomrule
\end{tabular}
\par\smallskip\footnotesize Each cell gives no-defense/joint-quarantine exposure.
\end{table}

The component results in Table~\ref{tab:components} compare recorded operating points. Abuse-only screening attains exposure 0.1436 but reduces clean NDCG by 0.1113. Instruction-only screening has a smaller clean loss while leaving exposure at 0.4169. Joint quarantine combines exposure 0.1489 with a reported mean clean change of $-0.0014$. The latter is the recorded aggregate; rounded per-retriever entries need not reproduce its last digit.

\begin{table}[t]
\caption{Component policies at their recorded operating points}
\label{tab:components}
\centering\small
\begin{tabular}{@{}lrr@{}}
\toprule
Policy & Exposure@10 & Clean $\Delta$NDCG@10\\
\midrule
Abuse only & .1436 & $-.1113$\\
Instruction only & .4169 & $-.0071$\\
Joint quarantine & .1489 & $-.0014$\\
\bottomrule
\end{tabular}
\par\smallskip\footnotesize Means across retrievers. Policies are not matched at a common false-quarantine rate; interaction effects are not isolated.
\end{table}

These are complete-policy comparisons. A matched $[r,a,i]$ model was not evaluated, and single-signal policies were not matched at a common false-quarantine rate. The observed trade-off cannot therefore be attributed to the interactions. A direct ablation must refit both feature sets on identical data and select separate thresholds under the same calibration constraint.

Continuous reranking is sensitive to the retriever. For the hybrid, exposure is 0.083 with a clean NDCG decrease of 0.062; under BM25, exposure remains 0.878. The two-stage penalty changes exposure by only 0.1--0.5 percentage points relative to quarantine while incurring a larger clean loss. These outcomes motivate direct quarantine as the primary reported policy, but they are also unmatched operating-point comparisons.

The instruction detector's held-out-category recall is 0.754, with calibration false quarantine 0.00076 at its reported operating point. Under hybrid retrieval, question-like sentiment requests reach post-quarantine exposure 0.500, followed by conversational requests at 0.476, research assistance at 0.346, task automation at 0.298, and business intelligence at 0.269. Smooth transitions from benign prose are also a challenge in TopicAttack~\cite{topicattack}. Adding the simulated source marker gives perfect calibration recall, but that marker encodes benchmark provenance and is excluded from the primary model.

\subsection{Exploratory downstream observations}
All 104 selected targets enter the undefended top five. After quarantine, 21 remain and 83 contexts no longer contain the target instruction. This is a retrieval observation and does not depend on whether an output is judged to execute the instruction.

Table~\ref{tab:generation} reports behavioral observations. Fourteen case pairs share queries, leaving 90 distinct clean queries. Across two models and two prompts, $4(90+104+104)=1,192$ responses cover clean, attacked, and filtered conditions. Expanding clean answers to associated cases produces $4\times3\times104=1,248$ judgments.

\begin{table}[t]
\caption{Exploratory generation observations by model and prompt}
\label{tab:generation}
\centering\small
\begin{tabular}{@{}llccc@{}}
\toprule
Model & Prompt & None & Joint & Clean correct\\
\midrule
Qwen3-4B & Conventional & 0/4 & 0/4 & 36/90\\
Qwen3-4B & Hardened & 0/4 & 0/4 & 34/90\\
Phi-4-mini & Conventional & 0/4 & 0/4 & 47/90\\
Phi-4-mini & Hardened & 0/4 & 0/4 & 47/90\\
\bottomrule
\end{tabular}
\par\smallskip\footnotesize None and Joint report confirmed/unresolved judgments among 104 attacked cases in each cell. Clean accuracy uses 90 distinct queries. Counts do not estimate a defense effect on attack success.
\end{table}

The endpoint confirms no hijacks, assigns failure to 1,200 judgments, and leaves 48 unresolved. Four underspecified monoalphabetic-substitution cases recur across model--prompt--condition cells. Both attacked conditions retain zero confirmed and four unresolved judgments per 104-case cell. Carrier removal does not resolve behavioral ambiguity or establish zero attack success.

Five provisional positives lacked the requested Caesar transformation; two misspelled claim labels lacked the requested automation output. Clean accuracy is 40.0\%/37.8\% for Qwen3-4B under conventional/hardened prompts and 52.2\% for Phi-4-mini under either prompt. The post hoc endpoint, repeated queries, and absence of confirmed undefended hijacks preclude an attack-success reduction estimate.

\section{Discussion and Reproducibility}
\subsection{Interpretation and remaining validation}
The measured benefit is a lower rate of target admission at small observed clean ranking cost. All page constructions were fixed before final evaluation. Adaptive attacks can exploit knowledge of a decision rule, and recent evaluations show that static attack suites can overstate protection~\cite{attackermoves}. Neither a construction-level holdout nor a defense-aware attack was evaluated here.

A matched-cost component experiment should compare $[r,a,i]$ with $[r,a,i,ar,ir]$ under identical splits and fitting procedures. A prespecified grid of calibration constraints would support threshold sensitivity, reporting test exposure, false quarantine, NDCG, and refill behavior without test-set retuning. Those comparisons remain unreported.

SciFact omits rendered pages, links, freshness, publisher reputation, and client-side content. Dense retrievers and independent construction families are needed for broader transfer. Dependence between queries sharing sources would require higher-level clustering from the original manifests.

\subsection{Layered deployment and residual errors}
Question-like instructions and copied evidence resemble legitimate content. Provenance or duplicate-cluster information may help, but remains untested. Source-level monitoring is also needed because average placement cost can hide repeated false quarantine of one publisher.

AttriGuard checks causal support for tool invocations, and Task Shield checks alignment with the authorized task~\cite{attriguard,taskshield}. Admitted candidates still require such downstream controls.

Blocker documents target answer availability, while opinion poisoning targets response stance~\cite{blocker,topicflip}. These harms are outside our task-displacement endpoint.

\subsection{Reproducibility status}
\label{sec:repro}
The reported run contains 31,500 query-level rows and 315,000 top-10 rows in 105 scenario--retriever--policy groups, each with 300 queries. Its accounting checks covered duplicate identifiers, absent or invalid scores, source joins, and target exposure in clean scenarios. These counts do not substitute for the underlying records.

Exact replay requires input hashes and folds, target manifests, normalization parameters, vectorizer vocabulary and tokenization, fitted coefficients, ranker settings, software versions, raw scores, and rankings before and after refill. Generation additionally requires model and quantization revisions, prompts, truncation code, responses, and endpoint labels. Calibration placement counts are also needed.

The accompanying sources recompile the manuscript, regenerate figures from reported aggregates, and reproduce the analytical bounds. Original candidate records and trained artifacts are unavailable in this revision. The package therefore cannot replay the experiments or supply the missing component and threshold analyses; no public anonymous experimental archive is claimed.

\section{Conclusion}
SearchInject applies a relevance-conditioned admission rule before retrieved pages enter model context. Across held-out queries and payload categories within four known construction families, the recorded policy reduces exposure by 83.2--83.7\%, while the largest absolute clean NDCG@10 change is 0.0018. The 0.5\% calibration constraint is empirical: observed clean test false quarantine is 0.59--0.65\%. Query-level analytical bounds account for repeated construction outcomes under an independent-query assumption.

Interaction effects, matched-cost comparisons, and downstream attack-success reduction remain unestablished. Resolving them requires original candidate records, additional calibrated comparisons, and a prospectively defined generation endpoint with confirmed undefended attacks.

\balance

\end{document}